# Guiding the Last Centimeter: Novel Anatomy-Aware Probe Servoing for Standardized Imaging Plane Navigation in Robotic Lung Ultrasound

Xihan Ma, Mingjie Zeng, Jeffrey C. Hill, Beatrice Hoffmann, Ziming Zhang, and Haichong K. Zhang, *Member, IEEE*

*Abstract*—Navigating the ultrasound (US) probe to the standardized imaging plane (SIP) for image acquisition is a critical but operator-dependent task in conventional freehand diagnostic US. Robotic US systems (RUSS) offer the potential to enhance imaging consistency by leveraging real-time US image feedback to optimize the probe pose, thereby reducing reliance on operator expertise. However, determining the proper approach to extracting generalizable features from the US images for probe pose adjustment remains challenging. In this work, we propose a SIP navigation framework for RUSS, exemplified in the context of robotic lung ultrasound (LUS). This framework facilitates automatic probe adjustment when in proximity to the SIP. This is achieved by explicitly extracting multiple anatomical features presented in real-time LUS images and performing non-patient-specific template matching to generate probe motion towards the SIP using image-based visual servoing (IBVS). The framework is further integrated with the active-sensing end-effector (A-SEE), a customized robot end-effector that leverages patient external body geometry to maintain optimal probe alignment with the contact surface, thus preserving US signal quality throughout the navigation. The proposed approach ensures procedural interpretability and inter-patient adaptability. Validation is conducted through anatomy-mimicking phantom and in-vivo evaluations involving five human subjects. The results show the framework's high navigating precision with the probe correctly located at the SIP for all cases, exhibiting positioning error of under 2 mm in translation and under 2 degrees in rotation. These results demonstrate the navigation process's capability to accommodate anatomical variations among patients.

*Note to Practitioners*—Compared with traditional freehand ultrasound (US) imaging, robotic ultrasound systems (RUSS) have the potential to largely standardize the US diagnosis outcome caused by varying operator expertise if an inter-patient consistent, automatic standardized imaging plane (SIP) navigation process is available. This paper presents a SIP navigation framework for lung US (LUS) examination, which recognizes anatomical landmarks from the US images and fine-tunes the pose of the US probe so that the landmarks are positioned in accordance with a non-patient-specific template image. The special end-effector, active-sensing end-effector (A-SEE), maintains the probe at an optimal orientation with respect to the body, allowing consistent-quality US images to be acquired throughout the navigation. Unlike previous works, our approach can navigate to complicated SIP containing multiple anatomies with interpretable robot arm motion. We verified our framework's ability to navigate the probe to the SIP with millimeter-level accuracy under phantom and human experiment settings. While preliminary results demonstrate the framework's efficacy in guiding the robotic LUS procedure, the performance of the system on other examinations (e.g., liver and thyroid US) involving soft tissues requires further validation. In the future, the framework can be applied in various US examinations by implementing specific anatomical feature detection modules.

This paragraph of the first footnote will contain the date on which you submitted your paper for review, which is populated by IEEE. This work was supported by the National Institutes of Health under Grants DP5 OD028162 and R01 DK133717. *(Corresponding author: Haichong K. Zhang.)*

This work involved human subjects or animals in its research. Approval of all ethical and experimental procedures and protocols was granted by the institutional research ethics committee at Worcester Polytechnic Institute, under Application No. IRB-21-0613.

Xihan Ma is with the Department of Robotics Engineering, Worcester Polytechnic Institute, Worcester, MA 01609 USA (e-mail: xma4@wpi.edu).

Mingjie Zeng is with the Department of Computer Science, Worcester Polytechnic Institute, Worcester, MA 01609 USA (e-mail: mzeng2@wpi.edu).

Jeffrey C. Hill is with the Department of Diagnostic Medical Sonography, School of Medical Imaging and Therapeutics, MCPHS University, Worcester, MA 01608 USA (e-mail: jeffrey.hill@mcphs.edu).

Beatrice Hoffmann is with the Department of Emergency Medicine, Beth Israel Deaconess Medical Center, Boston, MA 02215 USA (e-mail: bhoffma2@bidmc.harvard.edu).

Ziming Zhang is with the Department of Electrical and Computer Engineering, Worcester Polytechnic Institute, Worcester, MA 01609 USA (e-mail: zzhang15@wpi.edu).

Haichong K. Zhang is with the Department of Robotics Engineering, Worcester Polytechnic Institute, Worcester, MA 01609 USA, and also with the Department of Biomedical Engineering, Worcester Polytechnic Institute, Worcester, MA 01609 USA (e-mail: hzhang10@wpi.edu).

This letter has supplementary downloadable material available at XXX, provided by the authors.

Digital Object Identifier XXXXXX.

## I. INTRODUCTION

MEDICAL ultrasound (US) imaging is a well-established imaging tool and is employed for the diagnosis of countless diseases and pathologies in nearly every aspect of the human body. Common examples include abdominal organs, thyroid, liver, heart, ocular, musculoskeletal system, and increasingly also the lung [1]. In recent years, US image acquisition using robotic US systems (RUSS) has been extensively studied to address the limitations of traditional freehand-performed diagnostic US imaging [2], [3], [4], [5]. These limitations include having inconsistencies in exam outcomes because of varying operator expertise [6], being physically demanding to the sonographers due to the unergonomic probe-holding gesture [7], and the increased disease transmission risk stemming from physical interactions between the patient and the sonographer [8]. On the other hand, RUSS leverages the dexterity and accuracy of the robotic arm to manipulate the US probe mounted to the robot's end-effector. Hence, direct contact between the patient and the clinician can be avoided. In addition, its operations are often aided by automated sub-workflows such as preoperative imaging target localization [9], [10], probe contact force regulation [11], [12] and contact surface adaptation [13], [14], [15] to maximize procedural repeatability.

In freehand diagnostic US, the standardized imaging plane (SIP) is defined where the two-dimensional (2D) US image (B-mode US) slices the target structure from a certain orientation such that necessary diagnostic information can be acquired [16]. The SIP definitions are unique for different US examinations. Sonographers are trained to press the probe near the imaging target to receive quality echo signals, then fine-adjust the probe pose in six degree-of-freedom (DOF) to precisely localize the SIP by interpreting and manipulating anatomical landmarks observed in the US image stream. While the above process is critical to acquiring diagnostic US images, it is also the leading source for operator-induced outcome inconsistency. Therefore, RUSS needs to adhere to a coherent, interpretable, and effective SIP navigation workflow in an automated fashion to eliminate human error. To this end, real-time US images should be leveraged to establish a close-loop control scheme that brings the probe from the current imaging plane to the SIP. Meaningful spatiotemporal information needs to be extracted from the US images and converted to the desired probe motion.

Nevertheless, the development of a SIP navigation framework utilizing US image feedback remains an active research direction in the RUSS community, with challenges presented in the following aspects: *i)* the hard-to-interpret nature of US images: the planar images with limited retrievable information often suffer from noise and artifacts, as well as blurred tissue boundaries [17]. Hence, it is difficult to identify the useful content automatically, *ii)* the inter-patient variability which leads to different images of the same anatomy [16], making it hard for the SIP navigation to be generalizable across patients and extendable to a variety of clinical use cases, and *iii)* the vulnerability of the echo signal when the probe is sub-optimally aligned with the contact surface [18]. Therefore, the US image alone can be inadequate for a robust SIP navigation process.

### A. Related Works on SIP Navigation for RUSS

In the earlier attempts to implement SIP navigation in RUSS, Abolmaesumi *et al.* adopted the image-based visual servoing (IBVS) approach, where visual features are extracted from the US image and manipulated with appropriate probe motion. The approach was implemented for aortic artery imaging with the probe controlled in three DOF to maintain the centering of the artery section [19]. Mebarki *et al.* followed the same IBVS framework yet explored the use of image moments derived from the binary mask of the US-revealed anatomy as the visual feature [20]. This way, the IBVS approach can manipulate the probe in all six DOF while reaching the SIP. Nadeau *et al.* directly employed grayscale intensities in the image region-of-interest (ROI) as the visual feature to achieve computationally more efficient IBVS-based probe servoing for abdominal SIP navigation [21].

Nonetheless, the above works were only verified in phantom experiments where it was assumed that the target anatomy remains static and its complete shape can be captured in the image field-of-view (FOV). Under such conditions, the detection of the anatomies is relatively straightforward; thus, the IBVS-based approaches yielded satisfactory performance. In actual clinical scenarios, however, patient movements can easily interfere with the detection process, and the target anatomy may be partially observable in the limited FOV. In the pursuit of a robust SIP navigation framework when deployed on human patients, the US confidence map [17], which quantifies the per-pixel signal quality of the B-mode image, was leveraged for controlling the probe motion [22], [23]. With the confidence-map-driven control, the requirement for robust and precise anatomy delineation can be relaxed. This approach has been shown to work on actual patients. However, the SIP navigation was implicitly performed by optimizing the echo signal quality inside the ROI. In return, this lead to weakened anatomy awareness and caused less procedural interpretability.

Recent works have explored using artificial intelligence (AI) agents for the SIP navigation task. AI agents can be trained to recognize partially visible anatomies in US images with exceeded accuracy and speed over conventional image processing methods. The recognized US image can directly involve the robot's decision-making process, allowing an end-to-end pipeline to infer the optimal probe motion towards the SIP. For instance, Li *et al.* presented a simulation-verified reinforcement learning (RL) enabled framework for robotic lumbar spine SIP navigation [24]. The framework takes the current 2D B-mode image, and its confidence map as inputs and directly predicts the subsequent optimal probe motion toward the SIP. Huang *et al.* developed an imitation learning pipeline leveraging a convolutional neural network (CNN) based US image feature extractor to interpret the current US image and navigate through multiple SIPs of the carotid artery [25]. The pipeline was able to strictly follow the clinical protocol when tested on human patients. For the same robotic carotid artery imaging task, Bi *et al.* presented an RL-based SIP navigation agent that takes a queue of temporally ordered vessel segmentations to estimate the optimal probe movement [26]. Unlike [24], the agent was deployable on a physical robot for phantom imaging after being trained in simulation.

However, an AI agent capable of end-to-end SIP navigation for RUSS generally requires large-sized, professionally labeled training samples, which are difficult to obtain. The feasibility of deploying an AI agent trained in simulation on human subjects also requires further validation. Furthermore, compared to non-learning-based methods (e.g., IBVS-based ones), the actions taken by the AI agent are less explainable in terms of how the anatomical landmarks are utilized to manipulate the probe, which hinders the usage of AI-enabled approaches in the safety-critical medicine field.

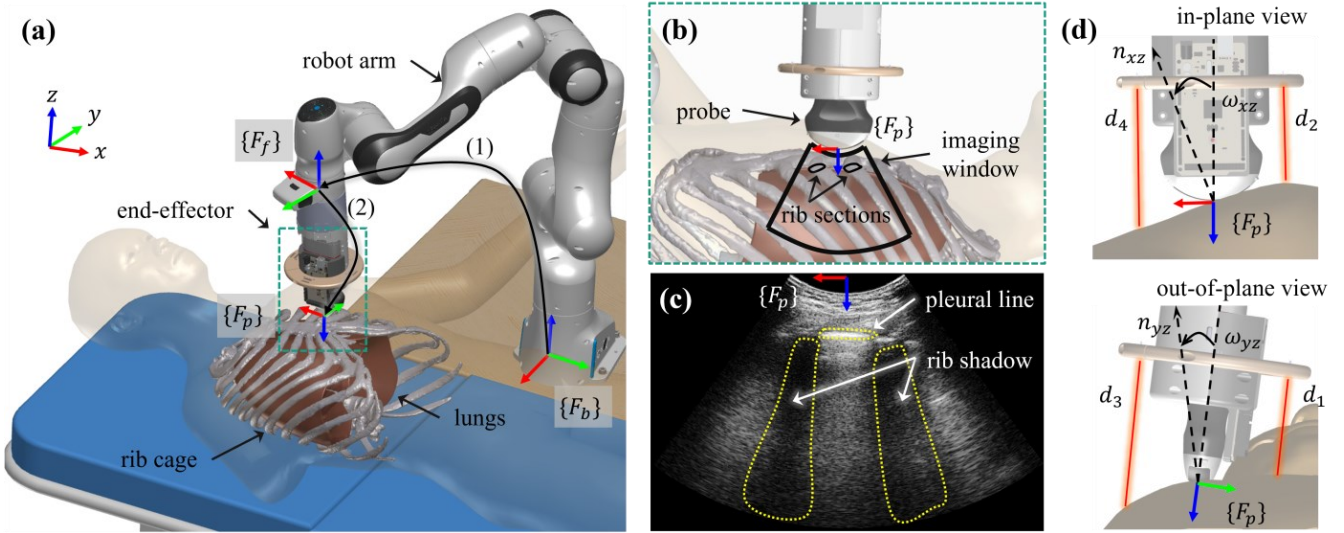

**Fig. 1.** RUSS-based LUS scene. (a) RUSS setup and coordinate conventions. $\{F_b\}$ stands for the robot base frame; $\{F_f\}$ stands for the flange frame at the robot's last link for attaching the end-effector. $\{F_p\}$ stands for US probe tip frame. (1) is the transformation from $\{F_b\}$ to $\{F_f\}$, denoted as $T_f^b$; (2) is the transformation from $\{F_f\}$ to $\{F_p\}$, denoted as $T_p^f$. (b) US probe placement to acquire the longitudinal intercostal view, i.e., SIP for LUS. (c) a representative US image acquired from the SIP. (d) probe in-plane view and out-of-plane view. The red beams stand for the distance measurement of the optical sensors.

Additionally, in most previous works, the SIP navigation primarily depends on US image feedback, which is likely to fail when the quality of the echo signal degrades due to improper alignment of the probe and the tissue. The definition of the SIP is rather simple, e.g., centering the vessel cross-section in the US FOV. In practice, however, the SIP may contain multiple anatomies at different locations of the image (e.g., lung [27] and fetal [16] US), warranting further enhancement on the navigation to handle nontrivial SIP definition. In summary, a SIP navigation framework that incorporates the strong US image interpretation power of the AI-based methods in accordance with the explainability of the classical approaches is desired for more complicated real-world scenarios.

### B. Contributions

To navigate SIP containing multiple anatomies in a real-world clinical setting, we propose an anatomy-aware SIP navigation framework for RUSS, combining AI-based US image visual feature extractor and IBVS-based probe servoing using the extracted feature. Lung US (LUS), an emerging examination for lung diseases, including COVID-19 diagnosis [27], [36], is selected as the target application since the SIP defined in LUS contains multiple (typically three) anatomical landmarks. The consistent identification of the SIP in LUS is critical to ensure accurate diagnosis across patients. In addition to the image-based probe servoing, our RUSS-integrated SIP navigation takes advantage of the robot end-effector's onboard sensing capability to continuously adjust the probe orientation according to the subject's body geometry, ensuring US image quality through the navigation process. The itemized contributions are as follows:

- A SIP navigation framework for RUSS-based LUS procedure is introduced. This real-time framework employs a CNN, trained using dataset created by ourselves on human subjects, to generate instance masks recognizing multiple anatomies from US images. These detected anatomies are further used to derive visual features for in-plane probe servoing (includes translational motion along the x-, and z-axis and rotational motion around the y-axis of $\{F_p\}$) under the IBVS scheme.

- We propose a novel concept called anatomical feature map (AFM) for robust IBVS using US images, from which generalizable visual features can be extracted for robust IBVS-based in-plane servoing. AFM represents a unique location mapping of multiple anatomical features from the current US image to a generic template image. It is aware of the geometric configuration among different anatomies and can accommodate inaccuracies in detecting anatomical landmarks.

- We integrate a customized robot end-effector, Active Sensing-End-Effector (A-SEE) [28] into the proposed SIP navigation framework. This integration allows us is to achieve scan surface adaptation, which optimizes the probe's in-plane rotation and out-of-plane rotation (rotation around the x-axis of $\{F_p\}$) simultaneously to track the normal direction of the body surface for optimal LUS image quality.

- The ability of performing SIP navigation for robotic LUS imaging using the proposed framework was verified on phantom and multiple human subjects, showing great potential for clinical usage.

To the best of our knowledge, this is the first work attempting to automate robotic LUS procedure by leveraging US image feedback. While LUS is used as an example application to demonstrate the clinical feasibility, our approach is potentially generalizable to other US examinations (will be discussed in Section V). The rest of this paper is organized as follows: Section II details the clinical background and technical components of the proposed SIP navigation framework; Section III describes the experiment conditions under which we validated the SIP navigation performance; Section IV presents the experiment results; Section V draws the conclusions, discusses limitations and future directions.

## II. MATERIALS AND METHODS

This section first formalizes the SIP navigation task using RUSS for LUS. The hardware components of our RUSS are then introduced. Lastly, the SIP navigation workflow is explained in detail.

### A. LUS SIP Navigation Problem Statement

The longitudinal intercostal view (see Fig. 1b) defined in the bedside lung ultrasound in emergency (BLUE) protocol [27] is adopted as the SIP in this paper. When placing the probe at the SIP, the US beam traveling through the rib space will reach the lung surface, creating the pleural line (PL) artifact in the image, whereas the beam hitting the ribs will be rapidly attenuated, leaving large, shadowed areas known as the rib shadow (RS). The acquisition of an LUS image from the SIP is depicted in Fig. 1b-c, where the PL is centered in the image and two chunks of RS are placed aside. PL and RS can be observed regardless of lung pathology, therefore are ideal anatomical landmarks for acquiring diagnostic lung US images. For LUS imaging, the probe must be oriented perpendicularly against the body to observe the PL and RS landmarks correctly. Previous works have demonstrated the initial landing of the probe near the SIP with a positioning error of under 2 cm [9], [29], allowing the capture of at least one anatomical landmark (either PL or RS) in the US image FOV. With the preliminaries above, the LUS SIP navigation task is described as follows: Assume a generic template SIP image, denoted as $\mathcal{T}$, is available. When pressing the probe on the subject's body with pose $T_p^b$ (see Fig. 1a), denote the received US image (query image) as $\mathcal{Q}(T_p^b)$. Under the proposition that the probe is already in the vicinity of the SIP after the initial landing, i.e., at least one anatomical landmark is visible, the goal is to adjust the probe pose based on the spatial configuration of the anatomies, such that the anatomies in $\mathcal{Q}$ (PL and RS) can be spatially aligned with the corresponding anatomies in $\mathcal{T}$.

### B. Technical Approach Overview

To accomplish the LUS SIP navigation task, we developed a RUSS which works in a realistic clinical environment. The system hardware setup is shown in Fig. 1a. A seven DOF robotic manipulator (Panda, Franka Emika, Germany) is mounted to the patient's bedside. The customized end-effector, A-SEE [28], which houses a wireless US probe (C3HD, Clarius, Canada) is attached to the robot. The probe is set to the manufacturer provided lung imaging mode for optimal brightness and contrast in LUS images with a default 10 cm depth penetration. The rigid body transformation from the flange frame $\{F_f\}$ to the probe frame $\{F_p\}$, $T_p^f$ (see Fig. 1a), is calibrated from the RUSS's CAD model. The calibration process does not take into account the error during manufacturing and assembly of the end-effector. However, we consider the final SIP navigation accuracy to be minimally affected by such errors as the navigation is a close-loop process leveraging US image feedback. Representing first-of-the-kind approach, A-SEE is used to sense the probe contact surface geometry, allowing autonomous probe normal positioning in both in-plane and out-of-plane. Its mechanism will be elaborated in section II E. Two workstation PCs are responsible for US image processing (at a rate of 30 Hz) and robot motion control (at a rate of 1000 Hz), respectively. Communication between the workstations is implemented through the Robot Operating System (ROS), achieving a round trip delay of approximately 6 ms.

During imaging, the probe is first located at the rough locations on the subject's chest area to allow the acquisition of US images. The LUS anatomical landmarks (PL and RS) are segmented from the image using a CNN-based segmentation network in real-time. Next, the semantic mask is instantiated before the AFM is constructed, where individual anatomies are matched with a generic LUS SIP template. Based on the geometric arrangement of the matched anatomies, visual features are extracted to calculate the desired in-plane probe velocity towards the SIP using an IBVS controller. In the interim, the required rotational velocity to maintain the probe in the normal direction of the probe contact surface is calculated using A-SEE. A velocity-based force controller actively regulates the probe contact force at a constant level for consistent acoustic coupling. A weighted sum of three velocities is performed to yield the final probe servoing motion.

### C. Probe Normal Positioning

As mentioned earlier, LUS requires the probe to be orthogonal to the contact surface while performing imaging. This essential requirement is satisfied using the A-SEE end-effector. Our previous work [28] has demonstrated the probe normal positioning feature using A-SEE which aligns the probe's long axis to the contact surface's normal vector in in-plane and out-of-plane separately. A-SEE employs four laser distance sensors (VL53L0X, STMicroelectronics, Switzerland) mounted on a sensor ring, providing four point-target measurements, noted as $d_1$, $d_2$, $d_3$, and $d_4$ as depicted in Fig. 1d. Upon orienting the probe perpendicularly to the body, the distance measurements of diagonally configured sensors should be the same. To achieve normal positioning, we control the two DOF probe angular velocities using a PD control law such that the differences in paired sensor readings are minimized. Symbolically, the angular velocity is calculated as follows:

$$\begin{bmatrix} \omega_{yz}^p \\ \omega_{xz}^p \end{bmatrix} = \begin{bmatrix} K_{pn} & K_{dn} & 0 & 0 \\ 0 & 0 & K_{pn} & K_{dn} \end{bmatrix} \begin{bmatrix} e_{13} \\ \dfrac{d}{dt}e_{13} \\ e_{24} \\ \dfrac{d}{dt}e_{24} \end{bmatrix} \qquad (1)$$

where $\omega_{yz}^p$ and $\omega_{xz}^p$ are the angular velocity around the x- and y-axis of $\{F_p\}$ respectively. The use of subscripts $xz$ and $yz$ suggest that the motion is happening in the in-plane and out-of-plane separately. $K_p$ and $K_d$ are empirically given gains, $e_{13} = d_3 - d_1$, $e_{24} = d_4 - d_2$, $\frac{d}{dt}(\cdot)$ computes the time derivative.

### D. Anatomical Landmark Segmentation

We implemented a LUS-Net (see Fig. 2) for simultaneous detection of PL and RS in real-time US images. LUS-Net consists of two U-Net [30] based sub-modules to learn the distinct features of PL and RS, respectively and generate the corresponding semantic masks for the two anatomical landmarks. To better capture the spatial relation between the anatomies, dilated convolution layers [31] were used in the networks' encoder path to increase the reception field, and attention gates [32] were added to the models' decoder path to capture the spatial context in the image.

LUS images were collected from a lung-mimicking phantom (COVID-19 Live Lung Ultrasound Simulator, CAE Healthcare™, USA) and human volunteers to train the LUS-Net (for later phantom and *in vivo* validation in section III). The phantom simulates signs for COVID-19, namely B-lines and white lung in US images (see Fig. 4f). Two models were trained separately for phantom and human LUS images due to differences in their image characteristics. 647 images were collected from the lung phantom. The human LUS image dataset consists of four sub-datasets with LUS images acquired from four male subjects. Each sub-dataset contains a slightly different number of images ranging from 92 to 159, resulting in a total number of 479 images. These images cover not only the LUS SIP view, but also views near the SIP. The images were labeled at pixel level under the instructions from clinical professionals.

Because the anatomical landmarks only account for a small area of the entire US image FOV, there is an imbalanced class problem when training the segmentation network. To mitigate this issue, we adopted a loss function that involves two terms to supervise the background and foreground (RS and PL) segmentation, respectively. The loss function $\mathcal{L}$ is defined as follows:

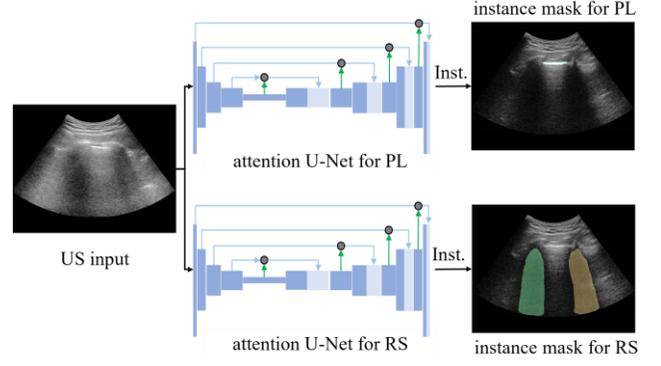

**Fig. 2.** LUS-Net schematics. LUS-Net consists of two attention U-Net modules connected in parallel, each takes the US input and predict the semantic mask before instantiating the masks. The blue blocks in the network represent convolutional layers. The circles represent attention gates. The blue arrows represent skip connections. The green arrows are the input signal to the attention gates. The outputs on the right show the color-coded instance masks of the PL and RS overlaid on the original US input.

$$\mathcal{L} = \alpha \cdot \mathcal{L}_{bce} + (1 - \alpha) \cdot \mathcal{L}_{dice} \qquad (2)$$

where $\alpha \in [0, 1]$ is a weighting factor, $\mathcal{L}_{bce}$ is the binary cross-entropy loss used to supervise the background class, calculated as:

$$\mathcal{L}_{bce} = -\frac{1}{M} * \sum_{i=1}^{M} y_i * \log \hat{y}_i + (1 - y_i) * \log(1 - \hat{y}_i) \qquad (3)$$

where $M$ is the number of pixels in the semantic mask, $y_i$ is the predicted label (either background or foreground), $\hat{y}_i$ is the ground truth label. $\mathcal{L}_{dice}$ is the Dice loss for supervising foreground classes. For each class (either PL or RS), $\mathcal{L}_{dice}$ is calculated as:

$$\mathcal{L}_{dice} = 1 - \underbrace{\frac{2 * \sum_{i=1}^{M} y_{f_i} * \widehat{y_{f_i}} + 1}{\sum_{i=1}^{M} y_{f_i}^2 + \sum_{i=1}^{M} \widehat{y_{f_i}}^2 + 1}}_{Dice\ index} \qquad (4)$$

where $\widehat{y_{f_i}}$ is the ground truth foreground label, $y_{f_i}$ is the predicted foreground label.

#### TABLE I.
LUS-NET SEGMENTATION PERFORMANCE

| LOO Dataset ID | PL Dice Score | RS Dice Score |
|---|---|---|
| 1 | 0.77 | 0.75 |
| 2 | 0.82 | 0.82 |
| 3 | 0.78 | 0.81 |
| 4 | 0.71 | 0.76 |

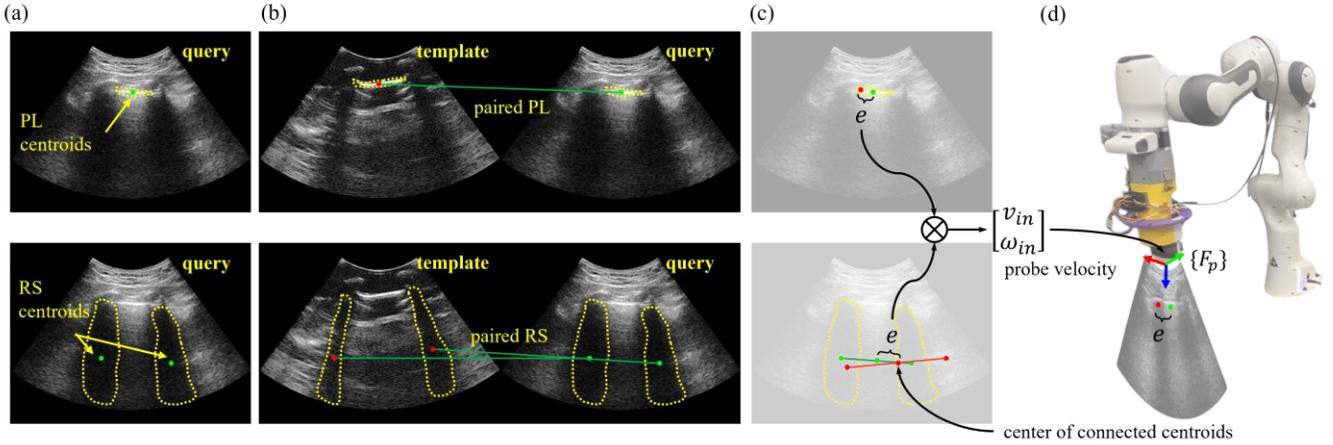

**Fig. 3.** 2-dimensional US image-based in-plane probe adjustment process. (a) Segmented PL (top) and RS (bottom) in the query image. The green dots represent the centroids of the instance mask for PL and RS. (b) AFM is constructed by pairing each anatomy in the query image with the corresponding one in the template image. (c) Illustration of the visual error derived from the misalignment of PL (top) and RS (bottom) with respect to the template image. The visual error is used to calculate the probe velocity based on the IBVS scheme. (d) Illustration of the visual error depicted under $\{F_p\}$.

When training the phantom LUS-Net model, 80% of the images were used for training and 20% were used for testing. To evaluate the segmentation performance on unseen human subjects, the human LUS-Net model was trained in a leave-one-out (LOO) fashion where three datasets were used for training, and one was used for testing. This process was repeated for every dataset. The Adam optimizer was used for both models with a batch size of 4 and an initial learning rate of 0.0002. Data augmentation, including random flipping and random rotation, was involved to accommodate the small training set. We achieved an accuracy of 71% for RS and 75% for PL in terms of the Dice index on the testing set for the phantom model. The LOO-based training result for the human model is shown in Table I. It can be concluded that the model was able to achieve an average segmentation accuracy of 77% for PL and 78% for RS in terms of the Dice index across all datasets. The performance variation across different training sets was 3.9% for PL and 3.0% for RS, showing good consistency and generalizability of the trained model.

For deployment, the model was trained using all four datasets to maximize the segmentation performance. The LUS-Net model was implemented using PyTorch. Both the training and the deployment of the model happened on the US image processing workstation equipped with a NVIDA RTX 3070 GPU.

### E. Anatomical Feature Map Generation

After detecting the LUS anatomies with the LUS-Net model, the next objective is to manipulate these anatomies in-plane so that each individual anatomy is positioned as desired. This is realized by IBVS-based probe pose control using AFM-derived visual features.

To generate AFM, we needed first to define the desired location for each anatomy at SIP. For this purpose, the template LUS image, $\mathcal{T}$, was collected from the lung phantom by a clinician, so that it is not bound to any specific subject. The SIP template was processed by the LUS-Net to obtain an instance mask of all PL and RS (see Fig 3b). For each category of the anatomy, the location of its instances is described by the first-order image moment (i.e., centroids) of the mask in 2D image space as point targets, denoted as $\hat{C} \in \mathbb{R}^{N \times 2}$, where $N$ is the total number of instances (see Fig. 3a). We use the first-order image moment as the location descriptor because it is less sensitive to morphological changes of the mask, therefore more robust to noise in the segmentation, partial visibility of the anatomy, and inter-patient variation in the shape of the anatomy. In the case of $N \geq 2$, we further link all centroids together and use the geometric center of the connected centroids as the location descriptor (see Fig. 3c). This supplementary operation is to achieve overall geometrical alignment of the anatomies across $\mathcal{T}$ (template SIP image) and $\mathcal{Q}$ (query image) despite the inevitable dissimilarity in their absolute pixel locations.

Second, given a query image $\mathcal{Q}$ with the same imaging FOV and pixel resolution as $\mathcal{T}$, the centroids of each category of detected anatomy are obtained in the same way and noted as $C \in \mathbb{R}^{n \times 2}$, where $n \leq N$ is the number of detected instances. Next, the centroids detected in $\mathcal{Q}$ need to be matched with those in $\mathcal{T}$ in order to calculate the probe pose error from the current imaging plane to the SIP. For a certain category of anatomy in $\mathcal{T}$, we can write all possible orders of instances as $\hat{L} \in \mathbb{R}^{P(N,n) \times 2}$, where $P(N,n)$ represents the permutation of $N$ elements taken $n$ at a time. We seek to find the optimal order of instances in the template, $\hat{L}^* \in \mathbb{R}^{n \times 2}$, such that the sum of image space distance between centroids of the corresponding instances is minimized. For each category of anatomy, the template matching can be formulated as the following optimization:

$$\hat{L}^* = \underset{\hat{L}}{\operatorname{argmin}} \sum_{j=1}^{n} \left\| C(j) - \hat{L}_i(j) \right\| \tag{5}$$

where $\hat{L}_i$ is the $i$-th permutation of instances, $\|\cdot\|$ computes the image space Cartesian distance of centroids. The above optimization was solved in brute-force by iterating through all permutations of instances. The cost for brute-force search is considered acceptable in the LUS case as the total number of anatomies is expected to be trivial (i.e., $\leq 3$). Similarly, the center of connected centroids is further obtained as the location descriptor for $N \geq 2$. This unique mapping of anatomical features across the template image $\mathcal{T}$ and the query image $\mathcal{Q}$ forms the AFM (see Fig. 3b).

### F. IBVS-based Probe In-plane Adjustment

With the AFM constructed, visual errors can be computed based on the misalignment of the paired anatomies. Consider a point target $C_i \in C$ with known three-dimensional coordinate under $\{F_p\}$, we define its visual error $\mathbf{e} = \begin{bmatrix} e_x & e_y & e_z \end{bmatrix}$ as the distance vector from the point target in $\mathcal{Q}$ to its desired location in $\mathcal{T}$. The visual error dynamics expressed in $\{F_p\}$ when applying a probe velocity $\begin{bmatrix} \mathbf{v_i} & \mathbf{\omega_i} \end{bmatrix}^T \in \mathbb{R}^6$ can be derived according to Fig 3d as:

$$\dot{\mathbf{e}} = \underbrace{\begin{bmatrix} -1 & 0 & 0 & 0 & -e_z & e_y \\ 0 & -1 & 0 & e_z & 0 & -e_x \\ 0 & 0 & -1 & -e_y & e_x & 0 \end{bmatrix}}_{L_m} \begin{bmatrix} v_i \\ \omega_i \end{bmatrix} \quad (6)$$

where $\dot{\mathbf{e}}$ is the time derivative of $\mathbf{e}$, $L_m$ is known as the interaction matrix. To impose an exponential decay on the visual error, an IBVS controller that controls the probe velocity can be constructed as:

$$\begin{bmatrix} \mathbf{v_i} \\ \mathbf{\omega_i} \end{bmatrix} = L_{m_i}^{\dagger} \lambda \mathbf{e_i} \quad (7)$$

where $\lambda$ is the empirically tunned control gain, $L_{m_i}^{\dagger}$ is the Moore-Penrose pseudo-inverse of the interaction matrix. Then, the probe velocity that manipulates all categories of features (i.e., PL and RS) is calculated as:

$$\begin{bmatrix} \mathbf{v_{in}^p} \\ \mathbf{\omega_{in}^p} \end{bmatrix} = \sum_{i=1}^{2} \gamma_i \cdot \begin{bmatrix} \mathbf{v_i} \\ \mathbf{\omega_i} \end{bmatrix} \quad (8)$$

where $\gamma_i$ is the weight adjusting the servoing priority for the $i$-th detected anatomy. In the case of LUS, PL is the primary diagnostic clue, hence the corresponding $\gamma$ value is larger.

Notice there is no guarantee that the spatial distance between paired anatomies (i.e., the visual error) can be eliminated to zero for all categories of anatomies due to inter-patient variation. The IBVS controller only attempts to minimize the distances with a user-specified order of priority, determined by $\gamma$. In case of oscillating probe motion due to residual errors, the image-based in-plane adjustment will be terminated when the error is below an empirically provided threshold.

### G. Contact Force Regulation

Given that the probe will be held perpendicular to the body, the contact force is regulated along the z-axis of $\{F_p\}$ through a velocity-based PD controller. The real-time contact force exerted at the probe tip is estimated by the robot joints' torque sensing at 1000 Hz. The controller adjusts the linear velocity along the z-axis. At time stamp $t$, the velocity is calculated as:

$$v_{fz}[t] = w \cdot K_{pf}(\widetilde{F_z} - F_z) + (1 - w) \cdot v_{fz}[t-1] \quad (9)$$

where $w \in [0, 1]$ is a weighting factor for smoothing the velocity profile, $F_z$ is the force exerted along the z-axis of the end-effector, $\widetilde{F_z}$ is the desired contact force which is empirically determined and set to be constant throughout the imaging.

### H. Probe Motion Fusion

Finally, we merge all the velocities profiles calculated above expressed under $\{F_p\}$ to yield the fused probe velocity $\begin{bmatrix} \mathbf{v^p} & \mathbf{\omega^p} \end{bmatrix}^T \in \mathbb{R}^6$:

$$\begin{bmatrix} \mathbf{v^p} \\ \mathbf{\omega^p} \end{bmatrix} = \underbrace{\begin{bmatrix} \mathbf{0_{3 \times 1}} \\ \omega_{yz}^p \\ 0 \\ 0 \end{bmatrix}}_{out-of-plane} + \underbrace{\begin{bmatrix} \mathbf{0_{3 \times 1}} \\ 0 \\ \omega_{xz}^p \\ 0 \end{bmatrix} + \begin{bmatrix} \mathbf{v_{in}^p} \\ \mathbf{\omega_{in}^p} \end{bmatrix}}_{in-plane} + \begin{bmatrix} 0 \\ 0 \\ v_{fz} \\ \mathbf{0_{3 \times 1}} \end{bmatrix} \quad (10)$$

where the first term computes the probe's out-of-plane motion, and the last three terms computes the probe's in-plane motion. The in-plane rotation incorporates both the motion resulting from A-SEE's normal positioning and the motion induced by the IBVS controller to align the anatomies. The merged probe velocity seeks to orient the probe to be perpendicular to the body in both in-plane and out-of-plane, maintain a constant contact force along the z-axis, then adjust the rest in-plane DOF to align the anatomical features with the template image for SIP navigation. The fused probe velocity is then transformed into the robot's base frame $\{F_b\}$ via:

$$\begin{bmatrix} \mathbf{v^b} \\ \mathbf{\omega^b} \end{bmatrix} = \begin{bmatrix} R_p^b & [\mathbf{p}] \\ \mathbf{0_{3 \times 3}} & R_p^b \end{bmatrix} \begin{bmatrix} \mathbf{v^p} \\ \mathbf{\omega^p} \end{bmatrix} \quad (11)$$

where $R_p^b \in SO(3)$ is the rotation part of $T_p^b$, $[\mathbf{p}]$ is the translation vector written in skew-symmetric matrix form. Next, the joint level velocity command $\dot{\mathbf{q}}$, which will be sent to the robot for execution is calculated as:

$$\dot{\mathbf{q}} = J(\mathbf{q})^{\dagger} \begin{bmatrix} \mathbf{v^b} \\ \mathbf{\omega^b} \end{bmatrix} \quad (12)$$

where $J(\mathbf{q})^{\dagger}$ is the Moore-Penrose pseudo-inverse of the robot's Jacobian.

## III. Experiment Setup

We evaluated the SIP navigation performance by testing our system on the aforementioned lung phantom, and on five male volunteers, respectively.

### A. Phantom Experiment Setup

The lung phantom was placed on the bed with US gel manually applied on the phantom surface (see Fig. 4a). The experiment procedure can be summarized in the three steps below.

*1) SIP identification:* The robot was first set to free-drag mode, where a sonographer dragged the US probe attached to the robot to scan the phantom and localize one SIP. Once the SIP was identified, the probe pose, as well as the US image, was recorded (referred to as the reference image and the reference probe pose). The reference image and the probe pose at the SIP served as the ground truth for later quantitative analysis.

*2) Apply deviation from SIP:* After collecting the reference image and the SIP probe pose, a slight translational offset was applied in the x-axis direction of $\{F_p\}$, deviating the probe from the SIP.

*3) SIP navigation:* Next, our SIP navigation framework was activated to bring the probe back to the initial pose (i.e., SIP pose) by leveraging the real-time US image feedback. The US image acquired preoperatively at a different SIP (i.e., different rib space) by the sonographer was used as the template image for navigation (Fig. 4e left). The LUS-Net trained on the phantom dataset was used to predict the anatomical landmarks.

We repeated *2)* to *3)* five times to study any inter-procedural variance. Note that to bring the probe to the initial pose, the amount of in-plane offset in *2)* can be no greater than $\frac{d_{rib}}{2}$, where $d_{rib}$ is the rib space measured from the center of consecutive rib cross-sections. This is because the SIP appears repeatedly between every two rib spaces as the probe travels along the subject's longitudinal axis. With an offset greater than $\frac{d_{rib}}{2}$, the probe may be brought to the "neighboring" SIP instead of the one initially identified by the sonographer. The rib space of the phantom is measured to be 35 mm, thus the in-plane offset amount was set to 17.5 mm. Nonetheless, such a proposition is only for the sake of quantitative evaluation. In clinical practice, being able to identify the anatomical landmarks at any rib space is considered as a successful SIP navigation. As we applied the translational offset, the probe also deviated from the optimal orientation due to change of the contact surface's normal vector. Therefore, the navigation framework needs to adjust the probe pose in both in-plane and out-of-plane to restore the SIP. A fixed duration of five seconds was employed for SIP navigation. Section IV will show that the five-second duration is sufficient for completing the navigation process.

During the SIP navigation, the contact force (i.e., the force measured along the z-axis under $\{F_p\}$) was continuously regulated to an empirically predetermined value adopted from the previous work [28] (3.5 N).

### B. Human Subject Experiment Setup

The same three-step procedure was followed for the human subject study. Five new male volunteers with varying body habitus were recruited whose LUS images were unseen to the pre-trained LUS-Net model, allowing us to assess the generalizability of the SIP framework. Among the subjects, three are over weighted, two are normal weighted, according to their body mass index (see Table II second column). US gel was applied to the subject's chest region before the experiment. Since the use of any specific subject's US image as the template may introduce bias, we continued using the SIP image obtained from the lung phantom (Fig. 4e left) as the template for navigation. The LUS-Net trained using all four human datasets was used for segmenting lung anatomies. At the reference probe pose, the half-rib spaces, $\frac{d_{rib}}{2}$, were measured to be 32 mm, 34 mm, 28 mm, 35 mm, and 30 mm for the subjects, respectively. Like the phantom study, the SIP navigation process was performed five times per subject, with a five second duration allowed for a single navigation process. The procedure was conducted on one side of the subject's anterior chest region. The desired contact force was set to 3.5 N for all five subjects.

### C. SIP Navigation Performance Evaluation Metrics

To evaluate the SIP navigation performance, we traced the similarity between the ground truth US image and the query US image during the navigation process. Three image similarity metrics were adopted, namely, the normalized cross-correlation (NCC) [33], the mutual information (MI) [34], and the sum of squared differences (SSD) of the pixel intensities. The SSD was further normalized (NSSD) for side-by-side display with the other two metrics:

$$NSSD = 1 - SSD / \max SSD \qquad (13)$$

While NSSD shows the per-pixel error between the current image and the ground truth image, US images usually contain significant speckle noise, artifacts, and random noise, making NSSD sensitive to non-critical information. Hence, NCC and MI, which capture the more clinical-relevant contexture information, were combined with NSSD for similarity comparison. In addition to the global US image similarity measure, the pixel-wise distance of paired LUS anatomies (PL and RS) between the query and the reference image was also recorded. The SIP navigation framework's ability to align the paired anatomical landmarks in AFM can be reflected by the convergence of the pixel-level separation curve. Moreover, we tracked the probe pose error to directly quantify the navigation accuracy. The probe pose error was obtained by computing the translational and rotational differences between the current probe pose and the reference probe pose. The probe contact force was measured at a frequency of 30 Hz to verify the efficacy of the contact force controller. The contact force error was defined as the root-mean-square of the desired contact force subtracted by the measured contact force.

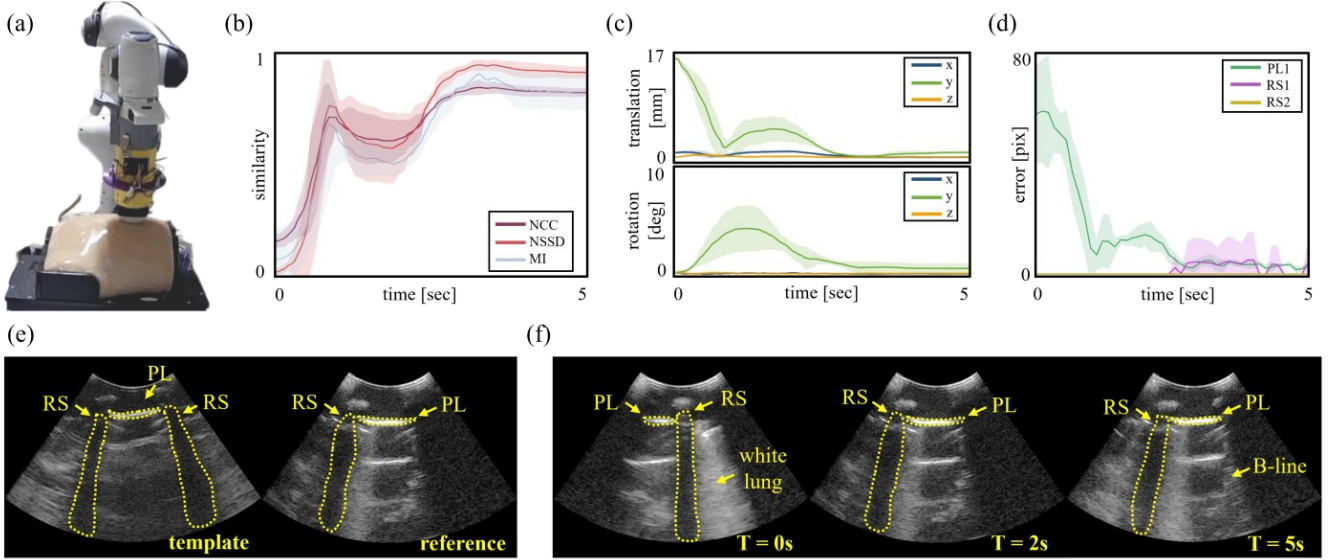

**Fig. 4.** Phantom experiment results. (a) Phantom experiment setup. (b) US image similarity measures between the query and the reference images (ground truth) throughout the SIP navigation process. The solid lines represent the mean similarity measure across five trials. The shaded area represents the standard deviation. (c) Probe pose error with respect to $\{F_b\}$ throughout the SIP navigation process. The top figure shows the translational error along the x-, y-, and z-axis. The bottom figure shows the rotational error about x-, y-, and z-axis. The solid lines represent the mean prose pose error across five trials. The shaded area represents the standard deviation. (d) Pixel distance between paired LUS anatomies. The absence of any curve during the time window indicates the corresponding anatomical feature was not detected. The solid lines represent the mean pixel distances across five trials. The shaded area represents the standard deviation. (e) The LUS image at the SIP (right), referred to as the reference image, and the template LUS image (left) acquired preoperatively at a different location from the reference image. Yellow dotted lines are automatic segmentation of the anatomical features. (f) Query images were recorded at different timestamps. Yellow dotted lines are automatic segmentation of the anatomical features. The phantom simulates COVID-19 signs including B-lines and while lung in US images.

## IV. RESULTS

This section presents results collected using the setup explained in the previous section. Readers can access raw experiment data, including recorded US images, robot trajectory, and contact force by sending requests to the authors.

### A. Phantom Validation on SIP Navigation Performance

The SIP navigation performance evaluated on the phantom is summarized in Fig. 4b-d. Fig. 4b shows the change in the similarity between the query and reference US image with respect to time. A steady increase in all three similarity measurements was observed over time. The similarity curves peaked after approximately three seconds, suggesting that the SIP can be localized in a short period of time. At the end of the SIP navigation, the mean US image similarity was improved by 68.7 % (NCC metric), 92.4 % (NSSD metric), and 77.0 % (MI metric), respectively. With the initial in-plane offset of 17.5 mm, the probe translational pose error was reduced to 1.06 ± 0.22 mm at the end of the SIP navigation (Fig. 4c). Due to the probe normal positioning feature, the rotational error first experienced a slight change in the out-of-plane direction (≤ 5 degrees) as the probe was sliding along the phantom surface, then rapidly converged to 0.55 ± 0.47 degrees. According to

Fig. 4d, while the distance of paired anatomical features witnessed an exponential decrease, the SIP navigation was primarily guided by aligning the PL features. Unlike the PL feature observable throughout the navigation process, the RS feature was less frequently detected (including at the SIP), and, hence, played a less important role in servoing the probe. This is because there is a distinct difference between the phantom and human rib structures in their acoustic properties: the acoustic wave cannot be sufficiently attenuated when passing through the rib in the phantom, making it challenging to identify the rib-caused acoustic shadow. Nevertheless, the purpose of assuring the RS feature can be involved in the probe pose optimization process has been fulfilled. Furthermore, when testing the same framework on human subjects, the RS features became easily identifiable, and thus were more frequently taken advantage of during the SIP navigation.

The US images captured at different timestamps in Fig 4f illustrate the dynamics of the SIP navigation process. Notably, the initial image at the zero-second mark exhibited considerable dissimilarity in the anatomical context compared to the reference image (Fig. 4e right). Yet, the final US image at the five-second mark showed a mostly identical alignment of the anatomical features compared to the reference image. The qualitative visualization further demonstrates the efficacy of the SIP navigation framework.

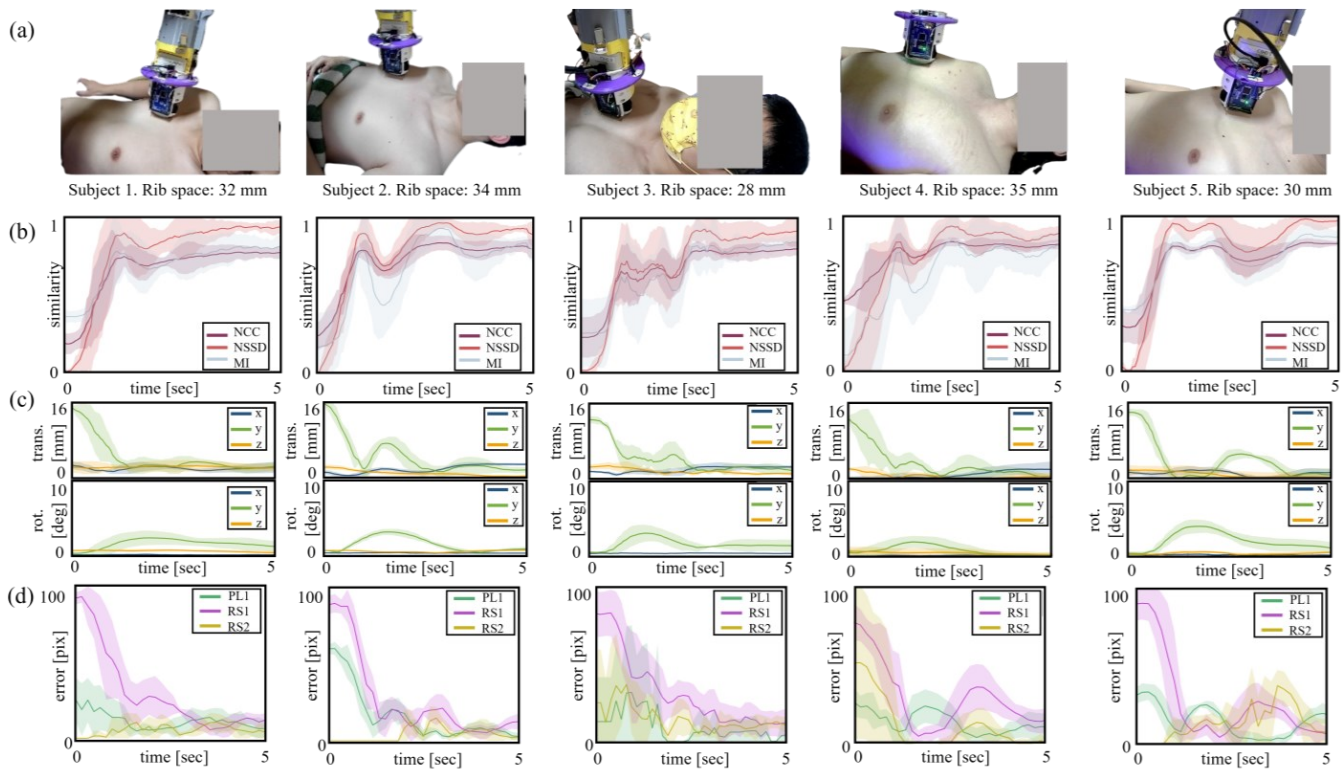

**Fig. 5.** Human subjects experiment with quantitative results. (a) Experiment setup for the five human subjects. (b) US image similarity measure between the query and reference images for the subjects (subjects 1 - 5 from left to right). The solid lines represent the mean similarity measure across five trials. The shaded area represents the standard deviation. (c) Probe pose error with respect to $\{F_b\}$ throughout the SIP navigation process for the five subjects. The top figure shows the translational error along the x-, y-, and z-axis. The bottom figure shows the rotational error about x-, y-, and z-axis. The solid lines represent the mean prose pose error across five trials. The shaded area represents the standard deviation. (d) Pixel distance between paired LUS anatomies for the five subjects. The absence of any curve during the time window indicates the corresponding anatomical feature was not detected. The solid lines represent the mean pixel distances across five trials. The shaded area represents the standard deviation.

## B. Human Subject Validation on SIP Navigation Performance

The results of the SIP navigation performance evaluation, conducted with human subjects, are shown in Fig. 5. Fig. 5b shows the US image similarity curve for the five subjects during the SIP navigation. Across all subjects, there was a notable and consistent increase in similarity between the query and the reference image throughout the navigation. On average, there was a growth of 52.2 % (NCC metric), 90.1 % (NSSD metric) and 56.8 % (MI metric) in the US image similarity for all subjects. The amount of increases in terms of the three metrics were seen in Table II third to fifth columns, respectively. By the end of the SIP navigation, the average probe translational and rotational error were $1.46 \pm 0.92$ mm and $0.81 \pm 0.57$ degrees, respectively. The probe pose errors for the five subjects separately can be found in the sixth and seventh column of Table II. The navigation process can be completed within four seconds for all subjects. Owing to the subjects' respiratory motion, both the image similarity curve and the probe pose error curve displayed minor fluctuations in the mean values, along with a slightly elevated standard deviation compared to the phantom experiment results. Yet, the SIP navigation framework was able to consistently servo the probe to the SIP with high accuracy, enabling the acquisition of diagnostic LUS images.

Fig. 5d indicates the active involvement of both PL and RS features in the navigation process. Similar to the results obtained in the phantom experiment result, the pixel distances of paired features demonstrated an exponential drop. However, the absolute value of the pixel errors tended to be larger in the human subject experiment. This discrepancy arises from the fact that the errors were calculated by comparing the query images from the subjects to a single generic LUS template obtained from the phantom. Consequently, it is expected to receive imperfect alignment of the anatomical features, leading to higher pixel-wise error. Nevertheless, given the minimized probe pose error at a sufficiently small value, the SIP navigation can be deemed effective.

Qualitative monitoring of the SIP navigation process for the subjects is shown in Fig. 6. There exist significant variations in the LUS images for different subjects. To summarize: i) the five subjects exhibited varying chest wall thickness, resulting in different axial positions of the anatomical landmarks, with subject 2 presenting the thickest chest wall; ii) While the US imaging parameters were set to be identical, the image contrast is noticeably different across the subjects, challenging the adaptability of the pre-trained LUS-Net model. Subject 3's images, as visually inspected, displayed the highest contrast; iii) the rib space distances ranged from 28 to 34 mm for the subjects, leading to distinct lateral positioning of RS and differing PL morphology. Despite all the inter-subject variations described above, the SIP navigation framework was able to restore the LUS image at the reference probe pose. This demonstrates the framework's robustness, enabling it to handle US images from previously unencountered human subjects effectively.

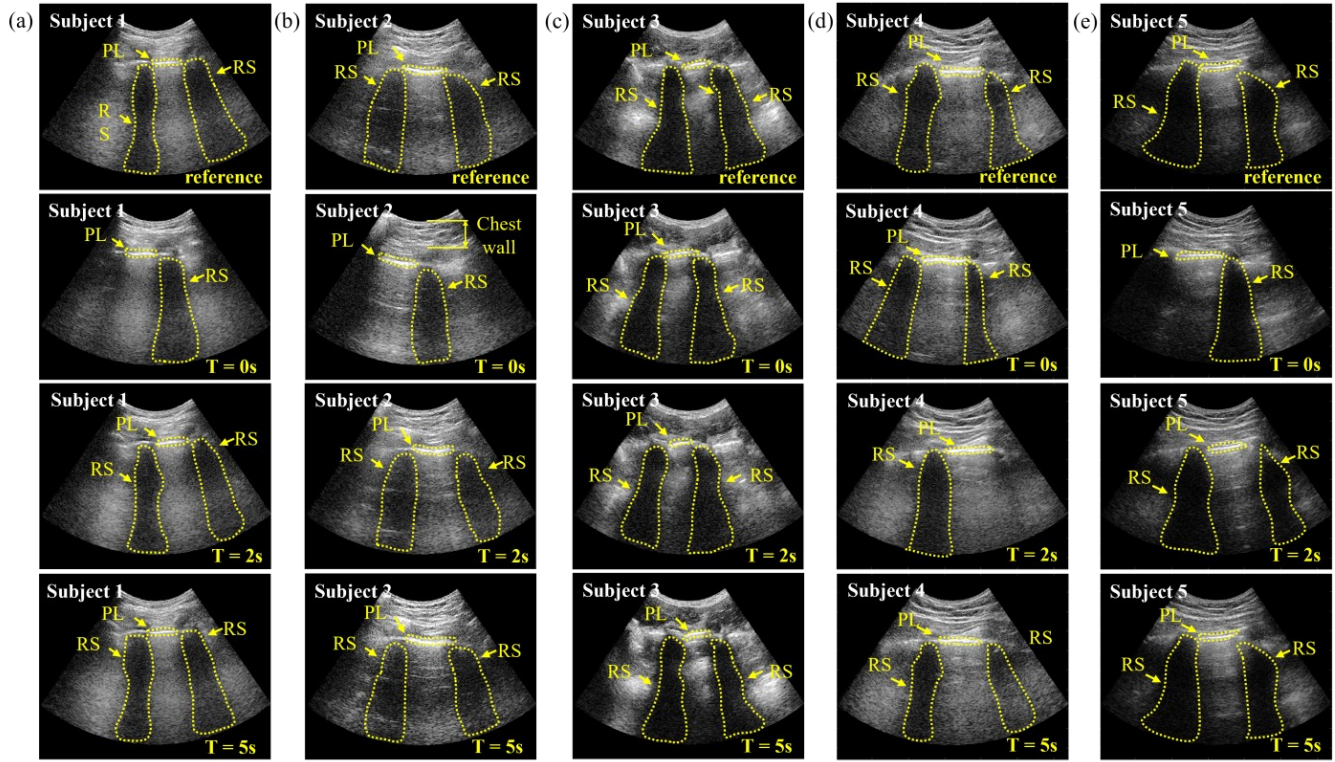

**Fig. 6.** Human subjects experiment qualitative results. (a-e) The LUS image at the SIP and the query images at different timestamps for subject 1 to subject 5. Yellow dotted lines are automatic segmentation of the anatomical features.

Compared to the phantom US images from Fig. 4e-f, the rib cross-section in the human LUS images is almost invisible due to acoustic attenuation, yielding a more identifiable rib shadow contour.

### C. Probe Contact Force Control Validation

The average contact force error when scanning the lung phantom was $0.59 \pm 0.12$ N. The average contact force error for scanning all five human subjects was $0.31 \pm 0.32$ N. The contact force errors for individual subjects were shown in the last column of Table II. The standard deviation of the measured force when scanning the human subjects was slightly higher than the phantom, likely influenced by variations in different trials and induced jittering attributed to respiration motion. Importantly, the contact force remained closely regulated near the desired value in both the phantom and human subject cases, underscoring the efficacy of the contact force controller. This was further validated during the human subject experiment, where none of the subjects reported any imaging-related discomfort due to excessive pressure. This observation highlights the appropriateness of the desired force value selection and affirms the effectiveness of the contact force control.

TABLE II.
SIP NAVIGATION PERFORMANCE ON HUMAN SUBJECTS

| Human Subject ID | Body-mass index (weight/height²) | Δ US Similarity (NCC) | Δ US Similarity (NSSD) | Δ US Similarity (MI) | Pose Error (trans.) | Pose Error (rot.) | Contact Force Error |
|---|---|---|---|---|---|---|---|
| 1 | 26.2 (82kg/177cm²) | 58.5% | 92.5% | 45.7% | $1.89 \pm 1.18$ mm | $1.16 \pm 0.76$ deg | $0.34 \pm 0.35$ N |
| 2 | 25.2 (78kg/175cm²) | 55.7% | 88.9% | 62.9% | $1.73 \pm 0.74$ mm | $0.59 \pm 0.27$ deg | $0.22 \pm 0.26$ N |
| 3 | 23.1 (63kg/165cm²) | 57.4% | 90.7% | 59.6% | $1.31 \pm 0.66$ mm | $1.11 \pm 0.78$ deg | $0.23 \pm 0.43$ N |
| 4 | 26.8 (85kg/178cm²) | 37.6% | 86.7% | 69.4% | $0.89 \pm 0.69$ mm | $0.28 \pm 0.31$ deg | $0.45 \pm 0.26$ N |
| 5 | 22.7 (68kg/173cm²) | 51.9% | 91.7% | 46.4% | $1.48 \pm 1.35$ mm | $0.93 \pm 0.71$ deg | $0.32 \pm 0.31$ N |

## V. DISCUSSION AND CONCLUSIONS

In this work, we introduce a SIP navigation framework for robotic LUS procedures, designed to guide the probe to the SIP after an initial positioning within 1-2 cm of the target (i.e., providing the last centimeter guidance). Our SIP navigation framework is aware of specific lung anatomies, thus, can provide consistent imaging outcomes across diverse patients. As an advantage over previously reported RUSS, A-SEE enables active probe orientation adjustment, ensuring echo signal quality throughout the SIP navigation. The in-vivo demonstration on the human subjects proves the framework's effectiveness under the presence of respiratory motion that would be encountered in a clinical scenario. It is worth noting that the proposed framework can be integrated into RUSS with ease to achieve a spectrum of autonomy levels, from semi-automated to fully autonomous, providing the "last centimeter" fine tuning of the US probe pose towards the SIP. Although LUS imaging is selected as an example application in this work, we have provided a generic formulation for the AFM generation and the IBVS-based probe servoing, which theoretically works for different anatomical structures of arbitrary numbers, thereby making the framework extendable to other US imaging tasks.

The preliminary results demonstrated that the proposed framework can precisely navigate the probe to the SIP with residual probe pose error of under 2 mm in translation and under 2 degrees in rotation. This degree of accuracy remains consistent across the phantom experiment and the human subject experiment. Upon completing the SIP navigation, diagnostic LUS images were successfully restored for the phantom and the three different human subjects. In all cases, a reference-query image similarity of over 90% (NSSD metric) underscores the framework's adaptability to subject variability. Moreover, subject safety can be guaranteed via the contact force controller as the fluctuation of the contact force can be maintained within 1 N.

An important limitation of this work pertains to the fact that while the five male subjects show variations in body habitus, they lacked any lung conditions at the time of being scanned. Nonetheless, since the phantom used in this study simulates COVID-19 signs, including B-lines and white lung (see Fig. 4f), the successful SIP navigation during the phantom study suggests a high likelihood of the system functioning normally when scanning human subjects with active lung symptoms. It is worth noting that none of the human subjects reported imaging-related discomfort due to the pressure of the probe's contact force. However, this may not be the case for subjects who are highly pain sensitive or with significantly larger body habitus that require excessive probe pressure. To address the above issues, ongoing efforts will involve expanding the participant pool to validate the system's feasibility for scanning a broader population, including both males and females with lung conditions. Additionally, future endeavors will aim to extend the navigation framework to other US imaging tasks, such as liver US [35], where the recognition of anatomical features presents unique challenges due to substantial anatomical complexity.

## ACKNOWLEDGMENT

The authors would like to thank Abhinav Palisetti for the assistance on data labeling and Zhaoyuan Ma for his insightful suggestions on algorithm development.

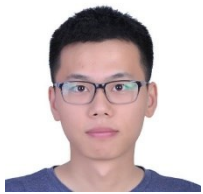
**Xihan Ma** received dual B.Eng. degrees in mechanical engineering and automation from Beijing University of Chemical Technology, Beijing, China, and mechatronics and robotics systems engineering from the University of Detroit Mercy, Detroit, Michigan, USA, in 2019. He received the M.Sc. degree in robotics engineering from Worcester Polytechnic Institute, Worcester, Massachusetts, USA, in 2021. He is currently pursuing Ph.D. degree in robotics engineering with the Department of Robotics Engineering, Worcester Polytechnic Institute, Worcester, Massachusetts, USA.

His research interests include medical imaging and medical robotics, with a focus on robotic ultrasound imaging and ultrasound guided intervention procedures.

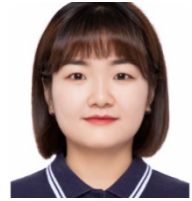
**Mingjie Zeng** received the B.S. degree in software engineering from Sichuan University, Chengdu, China, in 2020. She received the M.S. degree in computer science from Worcester Polytechnic Institute, Worcester, Massachusetts, USA, in 2023. She is pursuing a Ph.D. degree with Worcester Polytechnic Institute, Worcester, Massachusetts, USA. Her main research interests include deep learning and data mining, with a focus on neuroimage processing.

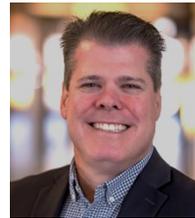
**Jeffrey C. Hill** is an Assistant Professor at the School of Medical Imaging and Therapeutics at Massachusetts College of Pharmacy and Health Sciences (MCPHS) University in Worcester, Massachusetts. Jeff earned his M.s.c. in Clinical Research at MCPHS University. He performed clinical research in echocardiography at Massachusetts General Hospital in Boston, MA, and UMass Medical School in Worcester, MA, for 11 years. He has been an educator for 13 years, teaching cardiovascular ultrasound principles and physics. Jeff's research interest includes the application of speckle-tracking strain imaging of the myocardium and pulmonary ultrasound. He is currently a co-investigator on an NIH-funded, tele-operative lung ultrasound study and is researching the application of *in-silico* models for assessing heart and vascular function and the aging aorta.

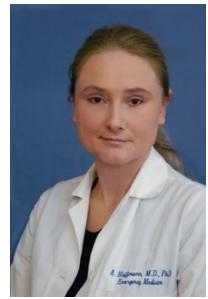
**Beatrice Hoffmann** is an Associate Professor in Emergency Medicine at Harvard Medical School and Division Director for Emergency Ultrasound in the Department of Emergency Medicine, at Beth Israel Deaconess Medical Center in Boston, MA. Her main research interests are point of care emergency medicine ultrasound applications, including lung ultrasound. Dr. Hoffmann received her M.D. from Heidelberg University in Germany and including her thesis on neurotransmitters in lung innervation including nitric oxide.

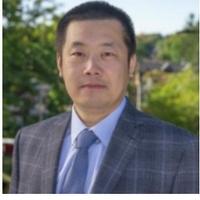

**Ziming Zhang** is an assistant professor at Worcester Polytechnic Institute (WPI). Before joining WPI he was a research scientist at Mitsubishi Electric Research Laboratories (MERL) in 2017-2019. Prior to that, he was a research assistant professor at Boston University in 2016-2017. Dr. Zhang received his PhD in 2013 from Oxford Brookes University, UK, under the supervision of Prof. Philip H. S. Torr. His research interests lie in computer vision and machine learning. He won the R&D 100 Award 2018.

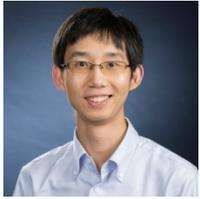

**Haichong K. Zhang** is an Assistant Professor in Biomedical Engineering and Robotics Engineering with an appointment in Computer Science at Worcester Polytechnic Institute (WPI). He is the founding director of the Medical Frontier Ultrasound Imaging and Robotic Instrumentation (Medical FUSION) Laboratory. The research in his lab focuses on the interface of medical imaging, sensing, and robotics, developing robotic-assisted imaging systems as well as image-guided robotic interventional platforms, where ultrasound and photoacoustic imaging are two key modalities to be investigated and integrated with robotics. Dr. Zhang received his B.S. and M.S. in Human Health Sciences from Kyoto University, Japan, and subsequently earned his M.S. and Ph.D. in Computer Science from Johns Hopkins University.